\documentclass[letterpaper]{article} 
\usepackage{aaai2026}  
\usepackage{times}  
\usepackage{helvet}  
\usepackage{courier}  
\usepackage[hyphens]{url}  
\usepackage{graphicx} 
\usepackage{natbib}  
\usepackage{caption} 
\usepackage{subcaption} 
\usepackage{algorithm}
\usepackage{algorithmic}
\usepackage{amsmath}  
\usepackage{tcolorbox}
\usepackage{booktabs} 
\usepackage{amssymb}  
\usepackage{multirow} 
\usepackage{array} 
\usepackage{newfloat}
\usepackage{listings}
\usepackage{amssymb}
\newcommand{\EquivFun}[2]{\mathsf{Equiv}(#1,#2)}
\newcommand{\PImp}[2]{\mathsf{Imply}_{P\rightarrow Q}(#1,#2)}
\newcommand{\QImp}[2]{\mathsf{Imply}_{Q\rightarrow P}(#1,#2)}
\newcommand{\Mutex}[2]{\mathsf{Mutex}(#1,#2)}
\newcommand{\IntersectFun}[2]{\mathsf{Intersect}(#1,#2)}
\newcommand{\Conflict}[2]{\mathsf{Conflict}(#1,#2)}
\newcommand{\Disagree}[2]{\mathsf{Disagree}(#1,#2)}
\newcommand{\IndependentRel}[2]{\mathsf{Indep}(#1,#2)}
\newcommand{\CompRed}[2]{\mathsf{CompRed}(#1,#2)}
\newcommand{\ContRed}[2]{\mathsf{ContRed}(#1,#2)}
\newcommand{\IntrConf}[2]{\mathsf{IntrConf}(#1,#2)}
\newcommand{\IntrDis}[2]{\mathsf{IntrDis}(#1,#2)}
\newcommand{\ImpConf}[2]{\mathsf{ImpConf}(#1,#2)}
\newcommand{\ImpDis}[2]{\mathsf{ImpDis}(#1,#2)}
\newcommand{\LocalConf}[2]{\mathsf{LocalConf}(#1,#2)}
\newcommand{\SpecPrior}{\mathsf{SpecPrior}}

\DeclareCaptionStyle{ruled}{labelfont=normalfont,labelsep=colon,strut=off} 
\floatstyle{ruled}
\newfloat{listing}{tb}{lst}{}
\floatname{listing}{Listing}
\title{Neuro-Symbolic Resolution of Recommendation Conflicts in Multimorbidity Clinical Guidelines}
\author{
    Shiyao Xie\textsuperscript{\rm 1,\rm 2},
    Jian Du\textsuperscript{\rm 1,\rm 2}\thanks{*Corresponding author.}
}
\affiliations{
    \textsuperscript{\rm 1}Peking University, National Institute of Health Data Science\\
    Beijing, CN\\
    \textsuperscript{\rm 2}Peking University Health Science Center, Institute of Medical Technology\\
    Beijing, CN
}

\usepackage{bibentry}

\begin{document}

\maketitle

\begin{abstract}
Clinical guidelines, typically developed by independent specialty societies, inherently exhibit substantial fragmentation, redundancy, and logical contradiction. These inconsistencies, particularly when applied to patients with multimorbidity, not only cause cognitive dissonance for clinicians but also introduce catastrophic noise into AI systems, rendering the standard Retrieval-Augmented Generation (RAG) system fragile and prone to hallucination. To address this fundamental reliability crisis, we introduce a Neuro-Symbolic framework that automates the detection of recommendation redundancies and conflicts. Our pipeline employs a multi-agent system to translate unstructured clinical natural language into rigorous symbolic logic language, which is then verified by a Satisfiability (SAT) solver. By formulating a hierarchical taxonomy of logical rule interactions, we identify a critical category termed Local Conflict—a decision conflict arising from the intersection of comorbidities. Evaluating our system on a curated benchmark of 12 authoritative SGLT2 inhibitor guidelines, we reveal that 90.6\% of conflicts are Local, a structural complexity that single-disease guidelines fail to address. While state-of-the-art LLMs fail in detecting these conflicts, our neuro-symbolic approach achieves an F1 score of 0.861. This work demonstrates that logical verification must precede retrieval, establishing a new technical standard for automated knowledge coordination in medical AI.
\end{abstract}

\begin{links}
    \link{Code and datasets are available at}{https://github.com/Shiyaoa/GuidelineCoordination}
\end{links}

\section{Introduction}
Retrieval-Augmented Generation (RAG) has emerged as the prevailing paradigm to mitigate the intrinsic limitations of Large Language Models (LLMs), such as hallucinations and knowledge obsolescence \cite{vladika2025facts, wu-etal-2025-assessing}, thereby enabling their widespread deployment in clinical decision-making \cite{liu2025improving,amugongo2025retrieval,yang2025retrieval}.By grounding model outputs in authoritative external sources like clinical guidelines, this approach aims to ensure safety and factual correctness \cite{zakka2024almanac,ong2024surgeryllm,kresevic2024optimization}.However, the efficacy of this paradigm is fundamentally limited by the logical integrity of the underlying knowledge base. In reality, contemporary guidelines exhibit substantial redundancy, inconsistency, and conflict across organizations, disease areas, and updates, particularly for patients with multimorbidity \cite{Blozik2013EpidemiologicalSF,Hoffmann2018TheIA,yaacoub2023exploring,tseng2025canadian}. When such conflicting documents are injected into the RAG context, even state-of-the-art LLMs can be misled, degrading performance and potentially amplifying clinical risk \cite{javadi2025evidence}. While some training-free \cite{Jin2024LongContextLM} or training-based \cite{Li2024RAGDDROR} methods attempt to mitigate noise, in the medical domain, conflicts among guidelines are not merely noise, they are the cause of logical inconsistency in AI systems and potential clinical errors. Consequently, merely patching the retrieval component is insufficient; there is an urgent need for Knowledge Governance targeting the guidelines themselves.

\begin{figure}[!ht]
\centering
\includegraphics[width=0.9\columnwidth]{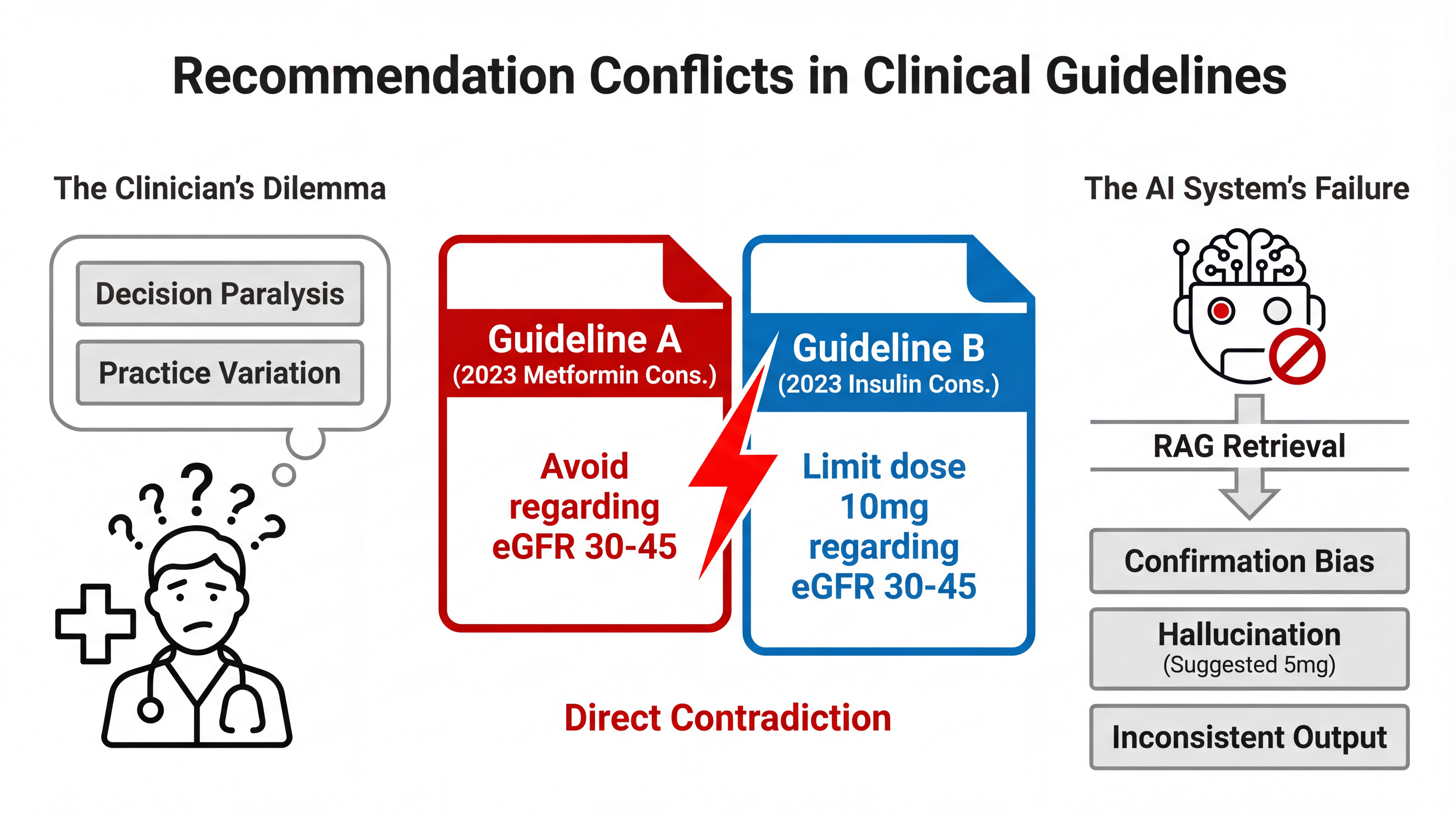}
\caption{Decision-making crisis caused by clinical guideline conflicts. Illustrated using two guidelines published in 2023 on Empagliflozin use in patients with eGFR 30-45, demonstrating direct contradictions.}
\label{fig:figure1}
\end{figure}

Conflicts among clinical guidelines primarily stem from organizational fragmentation \cite{tseng2025canadian}, variances in evidence interpretation \cite{Nagavci2025DefiningEO}, and a lack of standardized terminology \cite{Ghai2021EvaluationOC}. Specifically, these inconsistencies manifest as direct contradictions \cite{Raber2019TheRA}, temporal discrepancies, and most critically, multimorbidity conflicts arising from adverse interactions when single-disease guidelines are applied to multi-disease patients \cite{Dumbreck2015DrugdiseaseAD}. Analysis of primary cardiovascular disease prevention guidelines found that out of 124 recommendation clusters, only 44 clusters (35\%) included consistent recommendations, and merely 4 clusters (3\%) included highly consistent recommendations \cite{Bredehorst2025RecommendationsFT}. Such conflicts significantly increase polypharmacy risks, reduce patient adherence, and lead to medical resource wastage. Although the medical community has initiated harmonization efforts like Canada's C-CHANGE \cite{Jain2022CanadianCH}, existing approaches rely heavily on manual expert consensus. Given the exponential growth of medical literature, manual governance is no longer sustainable, creating an imperative for automated technical means to assist in this complex reasoning task.

From a computational perspective, guideline harmonization focuses on identifying and resolving redundancies and conflicts, which can be modeled as a Boolean Satisfiability (SAT) problem and solved using the Z3 theorem prover \cite{Moura2008Z3AE}. To address the probabilistic nature of LLMs in rigorous logical reasoning, Neuro-Symbolic AI has emerged as a promising solution. By leveraging LLMs as semantic parsers alongside external logical solvers, this paradigm has achieved remarkable progress in mathematical proving and formal logic tasks \cite{Pan2023LogicLMEL}. Despite this promise, neuro-symbolic methods have not yet been systematically explored for clinical guideline coordination. This is mainly due to two challenges: first, the extreme complexity of clinical natural language—replete with nested conditional clauses and ambiguity—makes high-fidelity Natural Language to Symbolic Language (NL2SL) translation difficult; second, the challenge of accurately modeling complex logical exclusions, particularly those involving multimorbidity.

To bridge these gaps, this paper presents an automated framework for clinical guideline conflict detection and harmonization. First, we designed a computable data model (schema) for clinical guidelines, formalizing unstructured recommendations into binary condition, action tuples. Second, drawing on Multi-Agent collaboration architectures, we developed a system to automate NL2SL translation and constructed a pipeline integrating a SAT solver to systematically verify guideline consistency. Finally, based on this pipeline, we evaluated the logical reasoning robustness of state-of-the-art LLMs under realistic RAG noise conditions. The main contributions of this paper are:

\begin{itemize}
\item We design a guideline-specific NL2SL schema that captures the nested conditional structure and action semantics of clinical recommendations.
\item We develop an automated neuro-symbolic pipeline that combines multi-agent LLM parsing with SAT-based verification to detect and categorize redundancies and conflicts, including multimorbidity-specific local conflicts.
\item We construct and open-source the first deep logical reasoning dataset dedicated to Guideline Harmonization, providing a new benchmark for knowledge governance methods for medical RAG systems.
\end{itemize}

\section{Related Work}
\subsection{Computable Clinical Guidelines} 
The transition from narrative guidelines to machine-executable formats has evolved from early rule-based systems to modern interoperable standards. Early formalisms like Arden Syntax pioneered knowledge representation but lacked support for shared data models \cite{Soares2021ACO}. Currently, Clinical Quality Language (CQL) serves as the standard for clinical decision support (CDS) and electronic clinical quality measures (eCQMs) \cite{Odigie2019FastHI,McClure2020IgnitingHD}. As an HL7 standard, CQL distinguishes itself through a data-model-agnostic architecture—compatible with FHIR, OMOP, and QDM—and a dual-representation system that offers both human-readable logic and a machine-executable Expression Logical Model (ELM) \cite{Brandt2020TowardCE}. This formalization enables the precise definition of cohort criteria and care gaps, leading to its widespread adoption by CMS and HEDIS for national quality programs \cite{Soares2021ACO}. However, despite CQL's expressive power, translating complex clinical prose into valid CQL libraries remains a labor-intensive bottleneck. It requires interdisciplinary experts to manually map ambiguous narrative exclusion criteria to strict logic and standardized terminologies \cite{Sittig2023ALF}, highlighting the urgent need for automated, neuro-symbolic translation mechanisms.
\subsection{Neuro-Symbolic Reasoning and Consistency Verification}
Neuro-symbolic frameworks employ LLMs as semantic parsers to translate natural language into formal representations such as  First-Order Logic or ASP for verification by deterministic solvers. Systems such as Logic-LM \cite{Pan2023LogicLMEL} and LINC \cite{Olausson2023LINCAN} demonstrate that offloading inference to external theorem provers significantly boosts performance on logical benchmarks. Frameworks like NL2FOL \cite{Lalwani2024AutoformalizingNL} and LELMA \cite{Mensfelt2024TowardsLS} leverage SMT solvers to explicitly detect logical fallacies and validity errors in LLM outputs.Recent research has critically extended this paradigm to normative conflict detection in high-stakes domains. In the legal field, \cite{Yadamsuren2025LLMAssistedFE} demonstrated that while pure LLMs struggle to identify statutory inconsistencies in tax codes, anchoring formalization in symbolic logic (Prolog) ensures deterministic detection. \cite{Mantravadi2025LegalWizAM} introduced LegalWiz, a multi-agent framework designed to stress-test legal RAG pipelines. Their work explicitly validates that unresolved contradictions in retrieved evidence lead to hallucinations, necessitating rigorous benchmarks with structured conflict types. While these legal frameworks offer valuable methodological parallels, clinical guidelines present a distinct challenge: unlike statutory contradictions, conflicts in multimorbidity often manifest as Local Conflicts—subtle logical intersections of conditional exclusions rather than direct negations—requiring the specialized hierarchical modeling and SAT-based verification proposed in this study.

\section{Methodology}
Our methodology transforms unstructured clinical guidelines into verifiable symbolic representations through a three-phase pipeline: Context Atomization, Neuro-Symbolic Formalization, and Logic-Based Verification.

\begin{figure*}[!ht]
\centering
\begin{subfigure}[t]{\textwidth}
\centering
\includegraphics[width=0.7\textwidth]{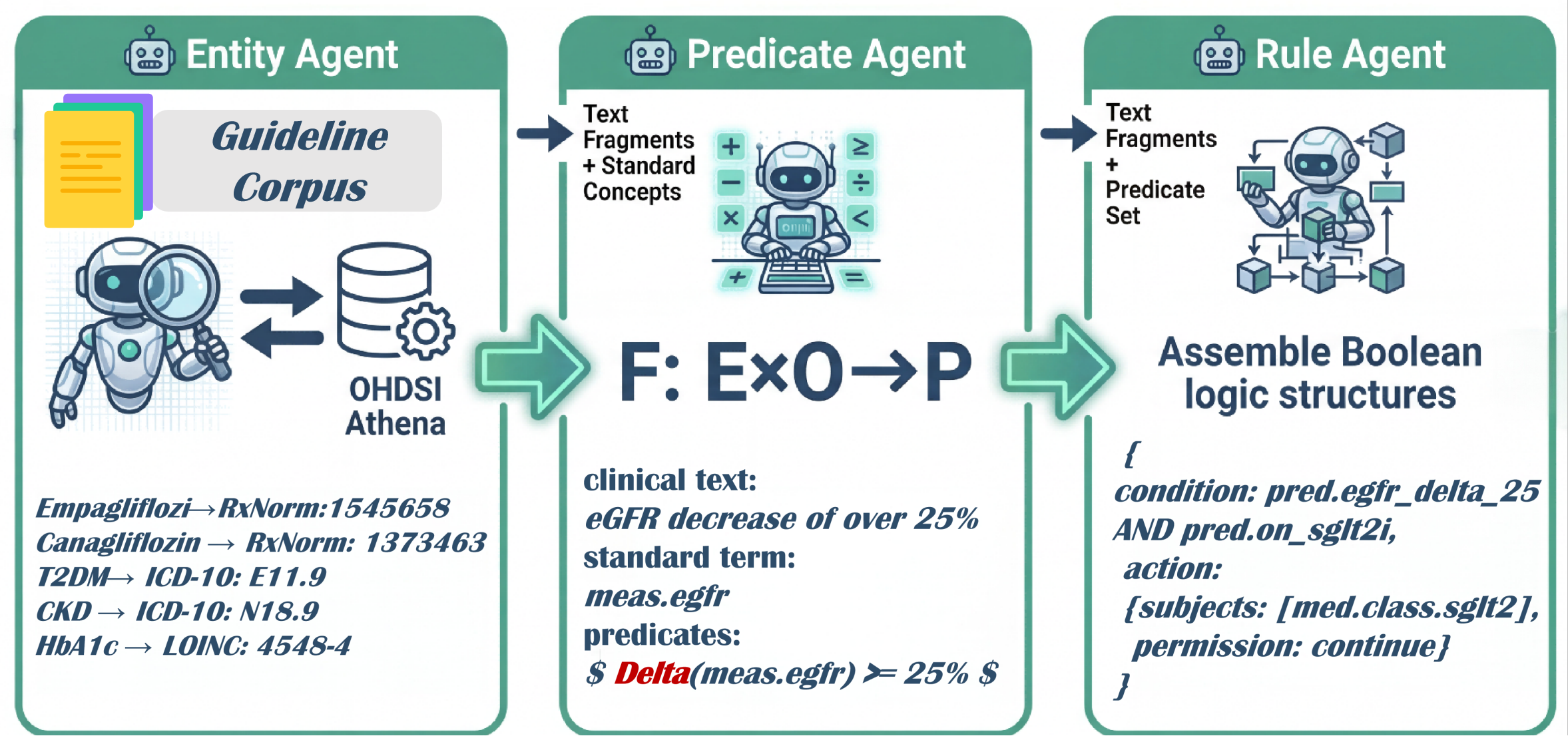}
\caption{Multi-Agent Formalization}
\label{fig:2a}
\end{subfigure}
\vspace{0.3cm}
\begin{subfigure}[t]{\textwidth}
\centering
\includegraphics[width=0.7\textwidth]{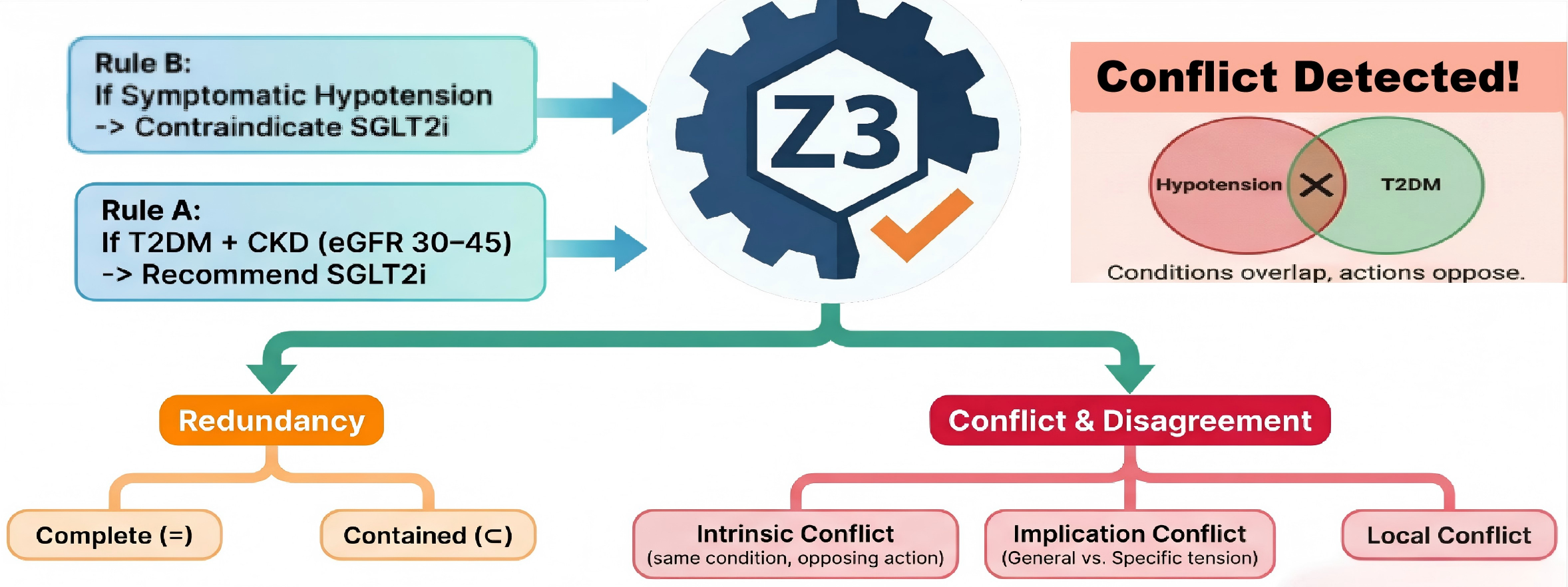}
\caption{Rule Relationship Analysis}
\label{fig:2b}
\end{subfigure}
\caption{The Neuro-Symbolic Pipeline for Clinical Guideline Formalization and Verification}
\label{fig:pipeline}
\end{figure*}

\subsection{Corpus Processing}
Prior to formalization, a \textit{Recommendation Extractor} functions as a pre-processor to stratify guideline content into four semantic categories: \textit{Risk Assessment}, \textit{Pharmacological Intervention}, \textit{Non-pharmacological Intervention}, and \textit{Other Opinions}. We specifically isolate \textbf{Pharmacological Interventions} for downstream processing using rigid syntactic inclusion criteria. To be retained, a recommendation must: (1) explicitly define both a \textit{target population} and a \textit{recommended action}; and (2) contain specific directive verbs (e.g., recommend, consider) and deontic modals (e.g., should, may). Following this extraction, we employ Locality-Sensitive Hashing (LSH) to group these recommendations into semantic clusters based on textual affinity

\subsection{Multi-Agent Formalization}
Within each cluster, a chained multi-agent system transforms natural language into executable symbolic logic ($\text{Terms} \rightarrow \text{Predicates} \rightarrow \text{Rules}$). This process addresses the semantic gap through three specialized agents:

\noindent\textbf{1. Entity Agent}
To ground ambiguous clinical text, this agent maps mentions to canonical codes in SNOMED CT, LOINC, and RxNorm via external retrieval tools. Crucially, this mapping adheres to \textbf{HL7 FHIR} standards, guaranteeing interoperability with real-world Hospital Information Systems (HIS).

\noindent\textbf{2. Predicate Agent}
A fundamental Semantic Gap exists between static \textit{Ontological Entities} (e.g., T2DM, insulin) and verifiable \textit{Logical Predicates} (e.g., Has T2DM, On insulin). Direct mapping leads to semantic ambiguity and restricts numerical reasoning. This agent bridges the gap between static entities and verifiable logic using a computable mapping $F: E \times O \rightarrow P$. It applies typed operators ($O$) to entities ($E$) to generate \textbf{Atomic Predicates} ($P$) compiled as SMT constraints (see Table~\ref{tab:operators} in Appendix~\ref{app:operators}):
\begin{itemize}
    \item \textbf{Existential Logic}: $O_{\text{exist}}(E)\rightarrow P_{\text{bool}}$.
    \item \textbf{Arithmetic Logic}: $O_{\text{arith}}(E)\rightarrow P_{\text{lra}}$.
    \item \textbf{Categorical Logic}: $O_{\text{cat}}(E)\rightarrow P_{\text{enum}}$.
\end{itemize}

\noindent\textbf{3. Rule Agent}
The final agent interprets semantic intent to assemble atomic predicates into \textbf{Compound Logical Structures} and maps directives to a standardized \textit{action vocabulary} (Appendix \ref{app:actions}). This produces a set of formal rules ready for solver-based verification.

\subsection{Rule Relationship Reasoning}
To ensure the clinical safety and consistency of the generated rules, we formulate the reasoning task as a \textbf{Satisfiability (SAT) problem}. By encoding the rules into the Z3 Solver (mathematical formulation detailed in \textbf{Appendix \ref{app:formula}}), we classify rule interactions into a hierarchical taxonomy comprising two coarse-grained categories and five fine-grained types:

\subsubsection{Category 1: Redundancy}
This category identifies rules that provide identical advice for overlapping populations, enabling knowledge base deduplication.
\begin{itemize}
    \item \textbf{Complete Redundancy}: Occurs when logical conditions are strictly equivalent ($C_1 \iff C_2$) and actions are identical.
    \item \textbf{Contained Redundancy}: Occurs when one rule's condition is strictly subsumed by another ($C_{\text{specific}} \subset C_{\text{general}}$), yet they dictate the same action. This implies that the specific rule is logically redundant within the scope of the general rule.
\end{itemize}

\subsubsection{Category 2: Conflict \& Disagreement}
This category captures logical incompatibilities where adherence to one rule violates another.
\begin{itemize}
    \item \textbf{Intrinsic Conflict}: The most direct inconsistency, where rules addressing the \textbf{exact same population} ($C_1 \iff C_2$) dictate opposing actions (\textit{Recommend} vs. \textit{Contraindicate}).
    \item \textbf{Implication Conflict}: Arises from a \textbf{General vs. Specific} tension ($C_{\text{specific}} \subset C_{\text{general}}$), where a subgroup guideline contradicts the general recommendation. Detecting this prevents the accidental application of broad guidelines to high-risk subgroups.
    \item \textbf{Local Conflict}: The critical source of multimorbidity issues. It occurs when rule conditions merely \textbf{intersect} ($C_1 \cap C_2 \neq \emptyset$) rather than encompass one another. When a patient simultaneously satisfies the conditions of multiple such rules, these rules yield conflicting decision recommendations. The conflict is local because it manifests exclusively in the patient subset suffering from both conditions simultaneously.
\end{itemize}

\section{Experiments}
Our experimental evaluation is designed to address two pivotal research questions:
RQ1: How accurately can our Multi-Agent system translate ambiguous, unstructured clinical natural language into the strict symbolic representations required by the solver?
RQ2: Is the proposed Neuro-Symbolic pipeline truly necessary? specifically, can standalone State-of-the-Art (SOTA) LLMs operating in a standard RAG setting achieve comparable reasoning performance without symbolic formalization?
\subsection{Experimental Setup}
\textbf{Dataset Construction.} We curated a benchmark corpus comprising \textbf{12 authoritative clinical guidelines} related to SGLT2 inhibitor (SGLT2i) therapy, sourced from diverse specialty societies within the Chinese Medical Association. We selected this domain as a representative testbed because the independent development of these guidelines naturally engenders a \textbf{full type of logical inconsistencies}, ranging from simple redundancy to complex intrinsic and implication conflicts. This heterogeneity provides a rigorous environment to evaluate the system's capability in logic formalization and relationship detection. Comprehensive details regarding the corpus composition are provided in \textbf{Appendix \ref{app:corpus}}.
Our pipeline utilizes DeepSeek-V3.1 \cite{deepseekai2024deepseekv3technicalreport} as the core semantic reasoning engine for all agentic tasks, while the Z3 SMT Solver serves as the symbolic verification backend. To answer RQ2, we benchmark our neuro-symbolic approach against top-tier proprietary and open-source LLMs: GPT-5 \cite{OpenAI2025GPT5SC}, Gemini-2.5-Pro \cite{comanici2025gemini}, Qwen-MAX \cite{qwen3max}, and DeepSeek-V3.1 (standalone). We evaluate their ability to detect redundancy and conflict relationships under varying noise levels, simulating real-world retrieval scenarios.
\subsection{Dataset Construction}
\textbf{Gold Standard Dataset}
To establish a rigorous benchmark, we constructed a high-quality dataset derived from the pipeline's output, subject to comprehensive human verification. Clinical experts first reviewed the candidate rule pairs to correct any misclassifications in the extraction or relation phases. To ensure dataset balance, we retained all instances of rare, critical relationships while down-sampling common ones. Additionally, to evaluate the models' capability to discern non-relationships, we incorporated negative samples consisting of rule pairs explicitly verified to have no logical interaction. 
\textbf{RAG Noise Dataset}
To evaluate model robustness under realistic retrieval setting, we constructed a Noisy Test Set using a  Graph-Based Strategy. We first modeled the entire rule base as a relationship graph, where nodes represent rules and edges represent verified logical relations. Within this graph, we identified Isolated Rules—nodes with a degree of zero, indicating no known logical interaction with any other rule in the corpus. We then generated test samples by injecting $k$ ($k \in [1,8]$) of these isolated rules into each verified base pair, followed by random shuffling to eliminate positional bias. This methodology guarantees that the injected noise functions purely as distractors without altering the fundamental logical relationship of the target pair.

\begin{table}[t]
\centering
\caption{Statistics of the Logical Benchmark Dataset}
\label{tab:dataset_stats}
\begin{tabular}{lcc}
\toprule
\textbf{Label Category} & \textbf{\#Pairs} & \textbf{Ratio} \\
\midrule
Conflict & 97 & 0.429 \\
\quad local\_conflict & 20 & 0.088 \\
\quad implication\_conflict\_or\_disagreement & 40 & 0.177 \\
\quad intrinsic\_conflict\_or\_disagreement & 37 & 0.164 \\
\midrule
Redundancy & 69 & 0.305 \\
\quad complete\_redundancy & 15 & 0.066 \\
\quad contained\_redundancy & 54 & 0.239 \\
\midrule
None & 60 & 0.265 \\
\midrule
\textbf{Total} & \textbf{226} & \textbf{1} \\
\bottomrule
\end{tabular}
\end{table}

\section{Results}
\subsection{Evaluation of Formalization Accuracy}
We first evaluated the foundational component of our pipeline—the NL-to-Symbolic translation—through rigorous manual audit. We posit that the validity of the downstream logical verification hinges entirely on the semantic fidelity of the LLM agents' formalization, as the Z3 solver's output is formally correct relative to its input.
An expert author with dual expertise in clinical medicine and data science manually audited 565 Rule objects generated by the pipeline. A rule was deemed Correct only under a Strict Exact-Match criterion: the formalized Predicate IDs and action (Permission, Subject) must be semantically identical to the original text provenance. Partial matches were classified as failures. The system achieved a Formalization Accuracy of 80.1\%. While this demonstrates strong zero-shot reasoning capabilities, 19.9\% of cases required correction. A granular breakdown of the 111 error cases reveals four primary failure modes:
\begin{itemize}
\item \textbf{Contextual Loss} (56.7\%, 63 cases): The most prevalent error involved missing implicit diagnostic context. For example, a recommendation extracted from a subsection titled Treatment for T2DM might fail to explicitly include \texttt{pred.has\_t2dm} in its predicate list, rendering the rule overly broad.
\item \textbf{Entity Grounding Failure} (19.6\%, 23 cases): This includes hallucinations or misclassifications of drug entities, such as confusing specific gliflozins or failing to map brand names to standard RxNorm terminologies.
\item \textbf{Predicate Distortion} (12.6\%, 14 cases): Logical errors such as hallucinating non-existent conditions, omitting critical exclusions, or inverting numerical operators.
\item \textbf{Action Alignment Error} (8.1\%, 9 cases): Discrepancies in the strength or direction of the recommendation, such as mapping should be considered to Recommend instead of Consider, or conflating Avoid with Contraindicate.
\end{itemize}
\subsection{Consistency Analysis of SGLT2i-Usage Guidelines}
Following the manual correction of formalization errors, we deployed the \mbox{Z3} SMT solver on the curated dataset to map the global logical landscape of SGLT2 inhibitor guidelines. This analysis revealed a stark contrast between general consensus and specific conflict.

\begin{table*}[t]
\centering
\caption{Representative Examples and Prevalence of Logical Interactions in SGLT2i Usage Guidelines}
\label{tab:logical_interactions}
\small
\begin{tabular}{p{1.5cm}p{1cm}p{2cm}p{2cm}>{\raggedright\arraybackslash}p{8cm}}
\toprule
\textbf{Relation Type} & \textbf{Count Pairs} & \textbf{Rule A} & \textbf{Rule B} & \textbf{Clinical explanations} \\
\midrule
Local Conflict & 2442 & If T2DM + CKD (eGFR 30--45) $\rightarrow$ Recommend SGLT2i. & If Symptomatic Hypotension $\rightarrow$ Contraindicate SGLT2i. & A patient with both conditions faces a clash between organ protection and immediate hemodynamic safety. The system flags this intersection, prompting a clinical priority setting. \\
\midrule
Implication Conflict & 115 & If eGFR $< 30$ $\rightarrow$ Contraindicate Metformin. & If eGFR $< 45$ $\rightarrow$ Avoid Metformin. & When a more specific condition triggers a stronger action than its broader, inclusive condition, creating contradictory guidance for the overlapping population. \\
\midrule
Intrinsic Conflict & 37 & If eGFR $< 15$ or Dialysis $\rightarrow$ Continue SGLT2i. & If eGFR $< 15$ or Dialysis $\rightarrow$ Stop SGLT2i. & Evidence Contradiction: Direct contradiction on the exact same population. Rare but critical, likely reflecting differences in guideline versions or evidence interpretation. \\
\midrule
Contained Redundancy & 57 & If T2DM + High CV Risk $\rightarrow$ Prioritize SGLT2i. & If T2DM + CKD (on ACEi/ARB) $\rightarrow$ Combine with SGLT2i. & Rule B is a subset of Rule A. The system validates that specific scenarios align with broader therapeutic principles. \\
\midrule
Complete Redundancy & 16 & If Severe Liver Impairment $\rightarrow$ Avoid SGLT2i. & If Severe Liver Impairment $\rightarrow$ Avoid SGLT2i. & Both rules enforce the exact same constraint. Identifying this allows for safe deduplication of the knowledge base. \\
\bottomrule
\end{tabular}
\end{table*}

The distribution of interaction types in Table 2 reveals that contradiction is rarely intrinsic, but frequently contextual. The overwhelming prevalence of Local Conflicts 2,442 pairs confirms that while guidelines are consistent for single diseases, they are structurally fragile for multimorbidity. The CKD vs. Hypotension example illustrates this perfectly: it is not a factual error, but a deadlock between competing clinical objectives long-term renal protection vs. immediate hemodynamic safety. This proves that RAG systems cannot simply retrieve documents; they require a Pathology Hierarchy to adjudicate such trade-offs.

The presence of 115 Implication Conflicts highlights the risk of semantic ambiguity, validating our need for a controlled action vocabulary. Conversely, true Intrinsic Conflicts are rare (37 pairs), suggesting that direct disputes over evidence are the exception rather than the norm.

\subsection{Baseline Comparison}

To answer RQ2 (Comparative Necessity), we benchmarked our fully automated pipeline against three state-of-the-art LLMs (DeepSeek-v3.1, Gemini-2.5-Pro, and GPT-5) operating in a standard RAG setting. It is important to note that the Ours metric reported here reflects the end-to-end performance of the automated agents without human correction. Despite the ~20\% extraction error rate above noted, our symbolic reasoning engine significantly outperforms pure neural baselines.

\begin{table}[t]
\centering
\caption{Performance Comparison on Logical Relation Detection}
\label{tab:baseline_comparison}
\small
\setlength{\tabcolsep}{4pt}
\resizebox{\columnwidth}{!}{%
\begin{tabular}{lcccc}
\toprule
\textbf{Relation} & \textbf{Model} & \textbf{Prec.} & \textbf{Rec.} & \textbf{F1} \\
\midrule
\multirow{5}{*}{Redundancy} & DeepSeek-v3.1 & 0.515 & 0.768 & 0.616 \\
 & Qwen-Max & 0.526 & 0.594 & 0.558 \\
 & Gemini-2.5-pro & 0.543 & 0.638 & 0.587 \\
 & GPT-5 & 0.507 & 0.493 & 0.5 \\
 & \textbf{Ours} & \textbf{0.689} & \textbf{0.775} & \textbf{0.729} \\
\midrule
\multirow{5}{*}{Conflict} & DeepSeek-v3.1 & 0.727 & 0.082 & 0.148 \\
 & Qwen-Max & 0.75 & 0.093 & 0.165 \\
 & Gemini-2.5-pro & 0.5 & 0.144 & 0.224 \\
 & GPT-5 & 0.75 & 0.186 & 0.298 \\
 & \textbf{Ours} & \textbf{0.826} & \textbf{0.898} & \textbf{0.861} \\
\bottomrule
\end{tabular}%
}
\end{table}

The results reveal a fundamental dichotomy in LLM capabilities:
Baselines achieve moderate performance on Redundancy tasks (F1: 0.50–0.62). This is expected, as redundancy often manifests as semantic similarity, which matches the pre-training objective of embedding models. However, ours still leads (F1: 0.729) because the Z3 solver eliminates pseudo-redundancies where text is similar but logical conditions differ slightly.
All baseline models failed catastrophically on Conflict detection, with Recall scores collapsing to 0.08–0.18. Even GPT-5 achieved an F1 of only 0.298. This indicates that LLMs struggle to distinguish topically related but contradictory information from topically related and consistent information. In contrast, our Neuro-Symbolic approach achieved an F1 of 0.861, demonstrating that mapping text to Boolean logic is the only reliable path for conflict detection.

\begin{figure}[!ht]
\centering
\includegraphics[width=0.9\columnwidth]{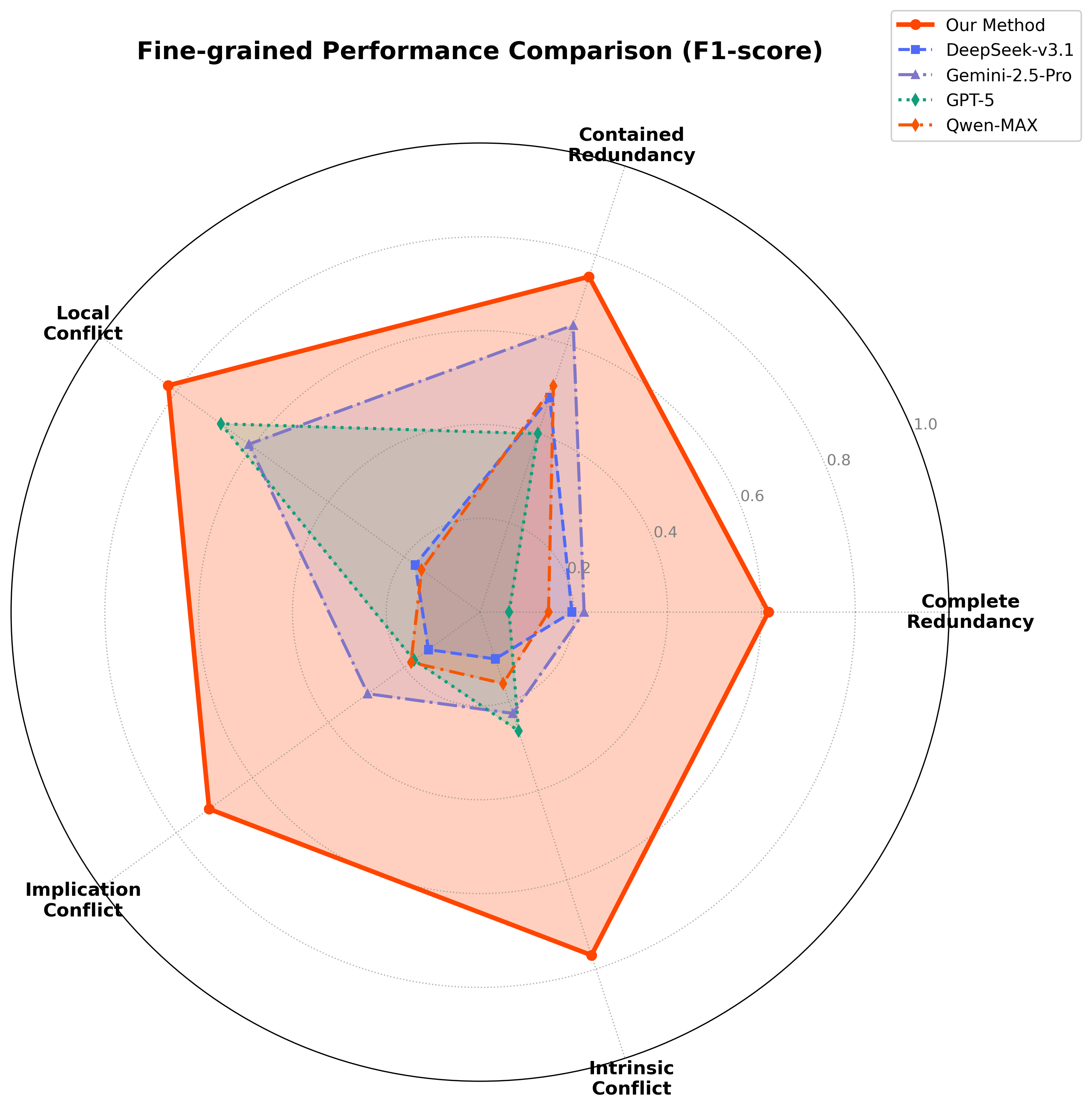}
\caption{Performance Comparison across Logical Sub-categories.}
\label{fig:fine_grained_performance}
\end{figure}

Figure~\ref{fig:fine_grained_performance} decomposes the F1-scores across the five fine-grained sub-types. Baselines perform worst on Local Conflict and Implication Conflict. These categories require understanding intersection and subset logic. Pure LLMs tend to classify these pairs as Related or Neutral because they lack the symbolic machinery to verify if the conditions overlap. Our method maintains a consistent hexagonal shape on the radar chart, indicating balanced performance across all logical types. The solver successfully bridges the gap, specifically excelling in Complete Redundancy and Intrinsic Conflict where logic is binary and absolute.

\begin{figure}[!ht]
\centering
\includegraphics[width=0.9\columnwidth]{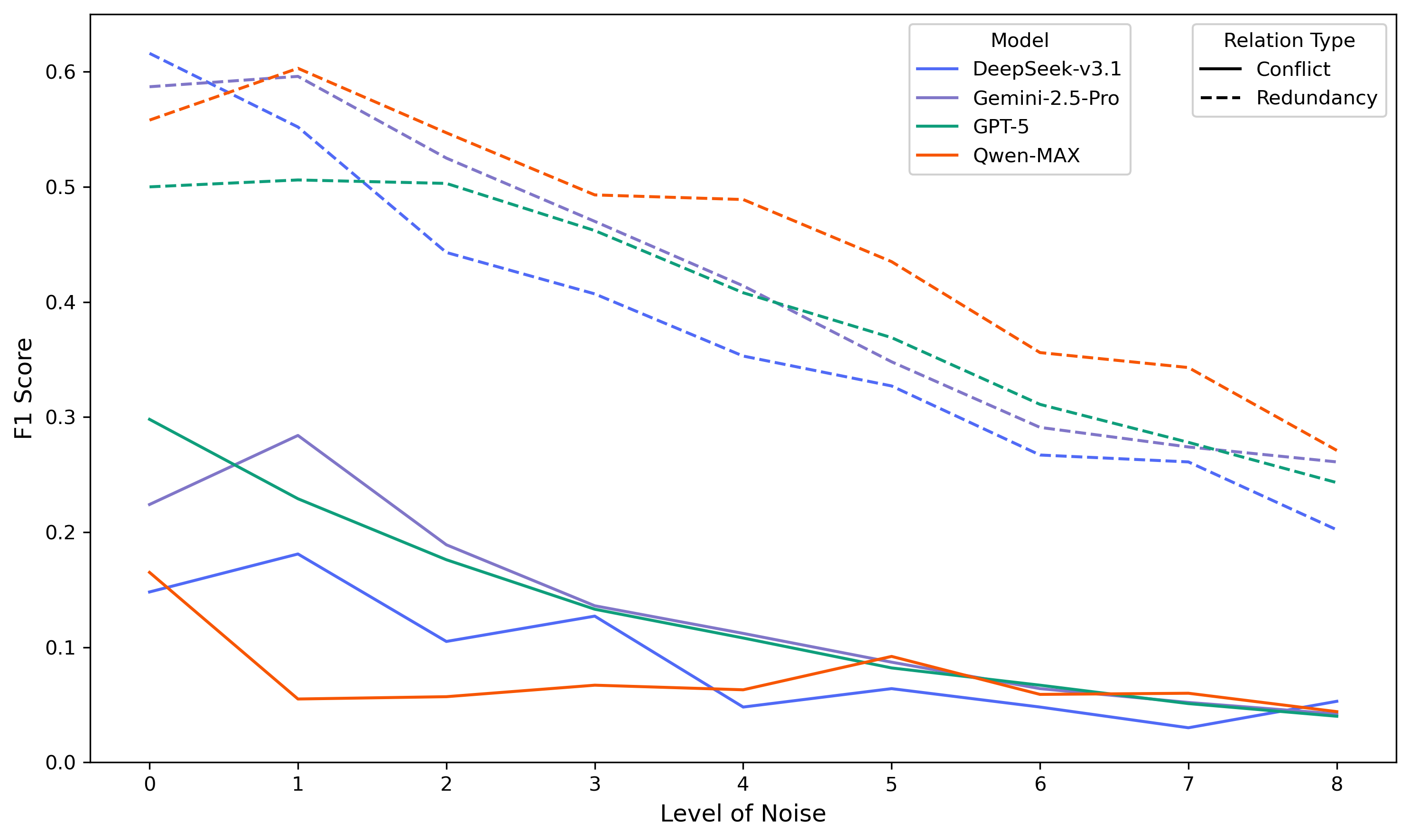}
\caption{Impact of RAG Retrieval Noise on Reasoning Performance.}
\label{fig:noise_comparison}
\end{figure}

Figure~\ref{fig:noise_comparison} illustrates the degradation of F1-scores as the number of noise rules increases from 0 to 8. All baseline models exhibit a sharp downward trend. As noise increases, the Signal-to-Noise Ratio in the context window drops, causing models to hallucinate relationships between unrelated rules or miss true conflicts due to attention dilution. These results confirm that pure RAG is insufficient for Guideline Harmonization. While LLMs are adequate for semantic retrieval, the Logic-Based Governance provided by our Neuro-Symbolic pipeline is indispensable for ensuring the safety and consistency of medical AI systems.

\section{Conclusion}
In summary, we highlighted a foundational yet often overlooked assumption in medical RAG: that clinical guidelines are logically consistent ground truth. We introduced a novel Neuro-Symbolic framework that synergizes the semantic flexibility of LLMs with the rigorous deducibility of SAT solvers to perform automated Knowledge Governance on clinical guidelines.
Our contributions are threefold. First, our methodology successfully formalizes ambiguous natural language into verifiable symbolic logic, achieving an 80\% accuracy in zero-shot translation. Second, our large-scale consistency analysis of SGLT2i usage guidelines uncovered a critical blind spot in modern medicine: 90.6\% of conflicts are Local, arising solely from multimorbidity intersections. This finding empirically demonstrates that single-disease guidelines are structurally inadequate for multi-disease patients, a complexity that standard vector-based retrieval cannot navigate. Third, our baseline comparisons reveal that even state-of-the-art LLMs suffer from a Reasoning Gap, failing catastrophically in conflict detection (Recall < 18\%) where our neuro-symbolic approach remains robust.
We conclude that logical verification must become a prerequisite for Medical RAG deployment. Merely retrieving information is insufficient; AI systems must possess the symbolic machinery to adjudicate the conflicting constraints of real-world pathology. Future work will extend this framework from conflict detection to automated harmonization, exploring how solver-generated proofs can guide LLMs to rewrite inconsistent guidelines into coherent, patient-centered decision pathways.

\bibliography{aaai2026}

\appendix
\section{SAT Modeling of Rule Relationships Using Z3-Solver}
\label{app:formula}

\subsection{Problem Formulation}
We model the problem of determining relationships between clinical rules as a \textbf{Satisfiability (SAT)} problem using the Z3 SMT solver. This approach provides a unified framework for reasoning about logical relationships between predicates, actions, and rules.

\subsection{Notation and Preliminaries}

Let $\mathcal{R} = \{R_1, R_2, \ldots, R_n\}$ be a set of clinical rules, where each rule $R_i$ is represented as:
\[
R_i: \quad C_i \Rightarrow A_i
\]
where:
\begin{itemize}
    \item $C_i$: a condition formula (Boolean combination of predicates over patient attributes)
    \item $A_i$: an action constraint formula (permissions, contraindications, dose restrictions, etc. on drugs/strategies)
\end{itemize}

Let $\mathcal{P} = \{P_1, P_2, \ldots, P_m\}$ be a set of predicates, where each predicate $P_j$ has a formal definition $\phi_j$ over a set of variables $\mathcal{V}$ representing the attributes of the patient like measurements, conditions, procedures.

\subsection{Predicate Relation Modeling}

\subsubsection{Variable Encoding}

For each predicate $P_i$ with formal definition $\phi_i$, we create a Boolean variable $p_i$ in Z3:
\[p_i = \text{Z3Expr}(\phi_i)\]
where $\text{Z3Expr}(\phi_i)$ converts the formal definition into a Z3 Boolean expression over variables in $\mathcal{V}$.

\subsubsection{Predicate Relation Definitions via SAT}

Given two predicates $P$ and $Q$ with Z3 expressions $p$ and $q$, we define their relationship using SAT queries:

\paragraph{Equivalence ($P \Leftrightarrow Q$)}
\[
\EquivFun{P}{Q} \Leftrightarrow \text{SAT}(p \neq q) = \text{UNSAT}
\]

\paragraph{Implication ($P \Rightarrow Q$)}
\[
\PImp{P}{Q} \Leftrightarrow \text{SAT}(p \land \neg q) = \text{UNSAT}
\]

\paragraph{Reverse Implication ($Q \Rightarrow P$)}
\[
\QImp{P}{Q} \Leftrightarrow \text{SAT}(q \land \neg p) = \text{UNSAT}
\]

\paragraph{Mutual Exclusion ($P \land Q$ is unsatisfiable)}
\[
\Mutex{P}{Q} \Leftrightarrow \text{SAT}(p \land q) = \text{UNSAT}
\]

\paragraph{Intersection (Satisfiable but neither implies the other)}
\begin{align*}
\IntersectFun{P}{Q} \Leftrightarrow {} & \text{SAT}(p \land q) = \text{SAT} \\
& \land \neg\PImp{P}{Q} \\
& \land \neg\QImp{P}{Q}
\end{align*}

\subsection{Action Relation Modeling}

\subsubsection{Action Representation}

An action $A$ is represented as:
\[
A = \{(s_1, \pi_1), (s_2, \pi_2), \ldots, (s_k, \pi_k)\}
\]
where:
\begin{itemize}
    \item $s_i \in \mathcal{S}$: a subject (drug/strategy) identifier
    \item $\pi_i \in \Pi$: a permission type like \texttt{allow}, \texttt{contraindicate}, \texttt{reduce\_dose}
\end{itemize}

\subsubsection{Action Variable Encoding}

For each action $A$ and each subject $s \in \mathcal{S}$, we define Boolean variables:
\begin{align}
\text{allow}_A(s) &: \text{subject } s \text{ is allowed by action } A \\
\text{prohibit}_A(s) &: \text{subject } s \text{ is prohibited by action } A
\end{align}

Constraints:
\[
\forall s \in \mathcal{S}: \neg(\text{allow}_A(s) \land \text{prohibit}_A(s))
\]

\subsubsection{Action Relation Definitions via SAT}

Given two actions $A$ and $B$:

\paragraph{Equivalence}
\begin{align*}
\EquivFun{A}{B} \Leftrightarrow {} & \forall s \in \mathcal{S}: \\
& \text{allow}_A(s) = \text{allow}_B(s) \\
& \land \text{prohibit}_A(s) = \text{prohibit}_B(s)
\end{align*}
This is checked by comparing the sets of subjects and permission values directly (no SAT query needed).

\paragraph{Conflict}
\begin{align*}
\Conflict{A}{B} \Leftrightarrow {} & \exists s \in \text{overlap}(A, B): \\
& \text{allow}_A(s) \land \text{prohibit}_B(s) \\
& \lor \text{prohibit}_A(s) \land \text{allow}_B(s)
\end{align*}
where $\text{overlap}(A, B) = \{s : s \in \text{subjects}(A) \cap \text{subjects}(B)\}$.

\paragraph{Disagreement}
\begin{align*}
\Disagree{A}{B} \Leftrightarrow {} & \exists s \in \text{overlap}(A, B): \\
& \text{permission}_A(s) \neq \text{permission}_B(s) \\
& \land \neg\Conflict{A}{B}
\end{align*}
This occurs when actions have overlapping subjects but different permission within the same category.

\paragraph{Independent}
\[
\IndependentRel{A}{B} \Leftrightarrow \text{overlap}(A, B) = \emptyset
\]

\subsection{Rule Relation Modeling}

\subsubsection{Rule Pair Analysis}

For a pair of rules $(R_a: C_a \Rightarrow A_a, R_b: C_b \Rightarrow A_b)$, we determine their relationship by combining predicate and action relations.

\subsubsection{Rule Relation Classification via SAT}

\paragraph{Complete Redundancy}
\begin{align*}
\CompRed{R_a}{R_b} \Leftrightarrow {} & \EquivFun{C_a}{C_b} \\
& \land \EquivFun{A_a}{A_b}
\end{align*}
\paragraph{Contained Redundancy}
\begin{align*}
\ContRed{R_a}{R_b} \Leftrightarrow {} & \PImp{C_a}{C_b} \\
& \lor \QImp{C_a}{C_b} \\
& \land \EquivFun{A_a}{A_b}
\end{align*}
\paragraph{Intrinsic Conflict}
\begin{align*}
\IntrConf{R_a}{R_b} \Leftrightarrow {} & \EquivFun{C_a}{C_b} \\
& \land \Conflict{A_a}{A_b}
\end{align*}
\paragraph{Intrinsic Disagreement}
\begin{align*}
\IntrDis{R_a}{R_b} \Leftrightarrow {} & \EquivFun{C_a}{C_b} \\
& \land \Disagree{A_a}{A_b}
\end{align*}
\paragraph{Implication Conflict}
\begin{align*}
\ImpConf{R_a}{R_b} \Leftrightarrow {} & \PImp{C_a}{C_b} \\
& \lor \QImp{C_a}{C_b} \\
& \land \Conflict{A_a}{A_b} \\
& \land \neg\SpecPrior
\end{align*}
\paragraph{Implication Disagreement}
\begin{align*}
\ImpDis{R_a}{R_b} \Leftrightarrow {} & \PImp{C_a}{C_b} \\
& \lor \QImp{C_a}{C_b} \\
& \land \Disagree{A_a}{A_b}
\end{align*}
\paragraph{Local Conflict}
\begin{align*}
\LocalConf{R_a}{R_b} \Leftrightarrow {} & \IntersectFun{C_a}{C_b} \\
& \land \Conflict{A_a}{A_b} \\
& \land \neg\SpecPrior
\end{align*}

\section{Study Corpus}
\label{app:corpus}

\subsection{The Chinese SGLT2i Multidisciplinary Guideline Dataset}
We curated a specialized corpus comprising \textbf{12 authoritative clinical guidelines} published by major branches of the \textbf{Chinese Medical Association (CMA)} and related professional societies, specifically the Diabetes Society, Society of Cardiology, Society of Nephrology, and the Chinese Geriatrics Society. The dataset focuses exclusively on \textbf{Sodium-Glucose Cotransporter-2 inhibitors (SGLT2i)}.

\subsection{Rationale for Corpus Selection: The SGLT2i Multimorbidity Nexus}
We selected SGLT2i therapy as the primary testbed because it perfectly exemplifies the \textbf{Organizational Fragmentation} and \textbf{Multimorbidity Conflicts} highlighted in our Introduction. The independent development of guidelines by distinct specialty societies creates a complex logical landscape that rigorously stress-tests our \textbf{Neuro-Symbolic Verification Pipeline}.

While both Cardiology and Nephrology guidelines recognize SGLT2i as a foundational therapy, they advocate for distinct drug combination paradigms based on their specialty focus. \textbf{Cardiology Guidelines}: Advocate for the Quadruple Therapy (SGLT2i + ARNI/ACEI + Beta-blocker + MRA) as the standard for Heart Failure. \textbf{Nephrology Guidelines}: Emphasize the Three Pillars of cardiorenal protection (SGLT2i + RAASi + MRA). For a patient with simultaneous Heart Failure and CKD (Cardiorenal Syndrome), these guidelines present \textbf{non-identical actionable sets}. The Nephrology does not explicitly mandate Beta-blockers, creating a discrepancy in the recommended set compared to the Cardiology. This tests the \textbf{Rule Agent's} ability to parse \textit{Conjunctions} and the \textbf{SMT Solver's} capacity to detect Inconsistent Detail when merging guidelines for multimorbid patients.

A primary source of inconsistency arises from how different specialties prioritize adverse risks. This creates contradictory stopping rules or monitoring logic, challenging our system's ability to handle complex exclusions.\textbf{Endocrinology}: Focuses heavily on \textit{Diabetic Ketoacidosis (DKA)}. Guidelines typically mandate discontinuation upon ketosis detection. \textbf{Cardiology}: Prioritizes hemodynamic stability. The primary constraint for suspension is symptomatic \textit{hypotension}, often tolerating metabolic fluctuations that might concern other specialists. \textbf{Nephrology}: Focuses on \textit{eGFR fluctuations} and \textit{Hyperkalemia}. Unlike simple binary triggers, renal guidelines contain complex conditional logic: an initial reversible dip in eGFR is expected and does not warrant discontinuation, whereas hyperkalemia requires managing the concomitant RAASi/MRA dosage rather than simply stopping SGLT2i. Detecting these nuanced constraints validates the \textbf{Entity Agent's} ontology mapping and the \textbf{Rule Agent's} synthesis of \textit{Compound Logical Structures}.

SGLT2i guidelines published at different times present conflicting  constraints due to rapidly updating evidence, particularly regarding eGFR thresholds. Older guidelines often set conservative exclusion criteria as eGFR $< 30$ mL/min/1.73m$^2$, while newer consensus documents lower this threshold to 20 or 25 or remove it for specific indications. This temporal discrepancy creates Implication Conflicts. It serves as a critical benchmark for our \textbf{Predicate Agent's} \textit{Arithmetic Operators} ($O_{\text{arith}}$), verifying whether the system can precisely model continuous numerical boundaries to identify outdated advice.

\section{Predicate Operator}
\label{app:operators}

The Predicate Agent employs a set of typed operators to transform ontological entities into verifiable logical predicates. Table~\ref{tab:operators} provides a comprehensive reference of all supported operators, their semantics, return types, Z3 mappings, and typical usage examples.

\begin{table*}[t]
\centering
\caption{Predicate Operator}
\label{tab:operators}
\small
\begin{tabular}{lcp{3cm}cp{4.5cm}}
\toprule
\textbf{Operator} & \textbf{Semantics} & \textbf{Return Type} & \textbf{Z3 Mapping} & \textbf{Typical Example} \\
\midrule
HAS & Disease diagnosis / existence & Bool & Bool & \texttt{HAS(cond.t2dm)} \\
ON & Medication status & Bool & Bool & \texttt{ON(med.insulin)} \\
HISTORY & Past history / historical events & Bool & Bool & \texttt{HISTORY(cond.stroke)} \\
ASSESS & Subjective assessment / clinical status & Bool & Bool & \texttt{ASSESS(Intolerance, med.statins) = True} \\
RISK & Risk rating & String & String & \texttt{RISK(cond.ascvd) = High} \\
STAGE & Staging / grading & Int/String & Int/String & \texttt{STAGE(cond.ckd) >= 3 or STAGE(cond.ckd) = B} \\
VALUE & Lab value / measurement & Real & Real & \texttt{VALUE(meas.egfr) < 45} \\
DURATION & Duration / length of time & Real & Real & \texttt{DURATION(med.insulin) >= 30} \\
DELTA & Change / delta & Real & Real & \texttt{DELTA(meas.creatinine) >= 50\%} \\
\bottomrule
\end{tabular}
\end{table*}

\section{Action Vocabulary}
\label{app:actions}

The Rule Agent maps clinical directives to a standardized action vocabulary with explicit partial ordering. This vocabulary enables precise conflict detection by establishing semantic relationships between different recommendation strengths. Table~\ref{tab:actions} provides a comprehensive reference of all supported action types, organized by category.

\begin{table*}[ht]
\centering
\caption{Action Vocabulary}
\label{tab:actions}
\small
\begin{tabular}{lp{6cm}p{4cm}}
\toprule
\textbf{Action Type} & \textbf{Semantics} & \textbf{Category} \\
\midrule
\multicolumn{3}{l}{\textit{Usage Control (with strength ordering)}} \\
\midrule
ALLOW & Allow use (optional alternative) & Usage Control \\
RECOMMEND & Recommend use (preferred choice) & Usage Control \\
REQUIRE & Require use (mandatory) & Usage Control \\
CONSIDER & Consider use & Usage Control \\
CAUTION & Use with caution (requires monitoring) & Usage Control \\
AVOID & Avoid use (better alternatives exist) & Usage Control \\
CONTRAINDICATE & Contraindicate (absolute prohibition) & Usage Control \\
\midrule
\multicolumn{3}{l}{\textit{Continuation Control}} \\
\midrule
CONTINUE & Continue current use & Continuation Control \\
STOP & Stop current use & Continuation Control \\
\midrule
\multicolumn{3}{l}{\textit{Dose Adjustment}} \\
\midrule
REDUCE\_DOSE & Reduce dose  & Dose Adjustment \\
INCREASE\_DOSE & Increase dose  & Dose Adjustment \\
START\_LOW\_DOSE & Start with low dose & Dose Adjustment \\
MAX\_DOSE\_LIMIT & Limit maximum dose & Dose Adjustment \\
TITRATE & Titrate adjustment based on efficacy/tolerance & Dose Adjustment \\
\bottomrule
\end{tabular}
\end{table*}

\end{document}